# The Power of Language: Understanding Sentiment Towards the Climate Emergency using Twitter Data

Arman Sarjou
*Departement of Computer Science*
*City, University of London*
London, United Kingdom
arman.sarjou.2@city.ac.uk

***Abstract*— Understanding how attitudes towards the Climate Emergency vary can hold the key to driving policy changes for effective action to mitigate climate related risk. The Oil and Gas industry account for a significant proportion of global emissions and so it could be speculated that there is a relationship between Crude Oil Futures and sentiment towards the Climate Emergency. Using Latent Dirichlet Allocation for Topic Modelling on a bespoke Twitter dataset, this study shows that it is possible to split the conversation surrounding the Climate Emergency into 3 distinct topics. Forecasting Crude Oil Futures using Seasonal AutoRegressive Integrated Moving Average Modelling gives promising results with a root mean squared error of 0.196 and 0.209 on the training and testing data respectively. Understanding variation in attitudes towards climate emergency provides inconclusive results which could be improved using spatial-temporal analysis methods such as Density Based Clustering (DBSCAN).**

***Keywords*— *Topic Modelling, t-Stochastic Neighbor Embedding, Seasonal AutoRegressive Integrated Moving Average Model, Sentiment Analysis***

## I. Introduction

According to the Intergovernmental Panel on Climate Change (IPCC), the impact of anthropogenic global warming on natural and human systems is already evident and in order to mitigate against further damage, it is imperative to reduce human-kind's collective greenhouse gas emissions [1]. The Climate Emergency is largely a middle-class problem, driven by globalization and mass consumerism [2], with 58% of total global energy consumption being carried out by the wealthiest 20% of individuals on the planet compared to just 4% by the poorest fifth of individuals on the planet [3]. As such, to mitigate against future climate related risk, policy changes will be required.

Understanding global attitudes towards the Climate Emergency will inform policymakers of the challenges that will be involved in enforcing meaningful action. Despite the large body of evidence that shows that the Climate Emergency has been caused by anthropogenic emissions, this is a controversial idea which is disputed within many political spheres [4]. This can make it difficult to challenge the pre-existing trends that have led humankind to this tipping point.

The burning of fossil fuels accounts for two-thirds of global greenhouse gas emissions [5]. As such, any effective action in reducing greenhouse gas emissions will pose an issue for the oil and gas industry and explains why there is often a reluctance from energy giants to commit to meaningful emissions reduction targets. This is most pertinently seen by ExxonMobil's long ties with climate denialist groups [6].

This study aims to investigate attitudes towards the Climate Emergency with an aim to direct future strategy in marketing to gain public support. In addition, this study aims to investigate the relationship between Crude Oil Futures (COF) and sentiment towards the Climate Emergency.

## II. Analyitical Questions and Data

### *A. Data*

Twitter has distinguished itself as a platform that allows the dissemination of real time news, political discussion and personal opinion. Every hour, over one million messages, with maximum length 280 characters, are posted to be viewed by users [7]. The brevity and volume of the messages makes Twitter a strong data source to gauge opinion among Twitter Users. This is suitable for this problem as Twitter users will be within the global middle/upper class which are among the highest greenhouse gas emitters [2].

The Twitter dataset used in this study was web scraped over 10 days from 19/10/2020-29/10/2020 using the Twitter API. Approximately 100,000 tweets were collected from the following search queries:

- Climate Change
- Climate Emergency
- Global Warming
- Our Planet
- Paris Agreement
- Environmental

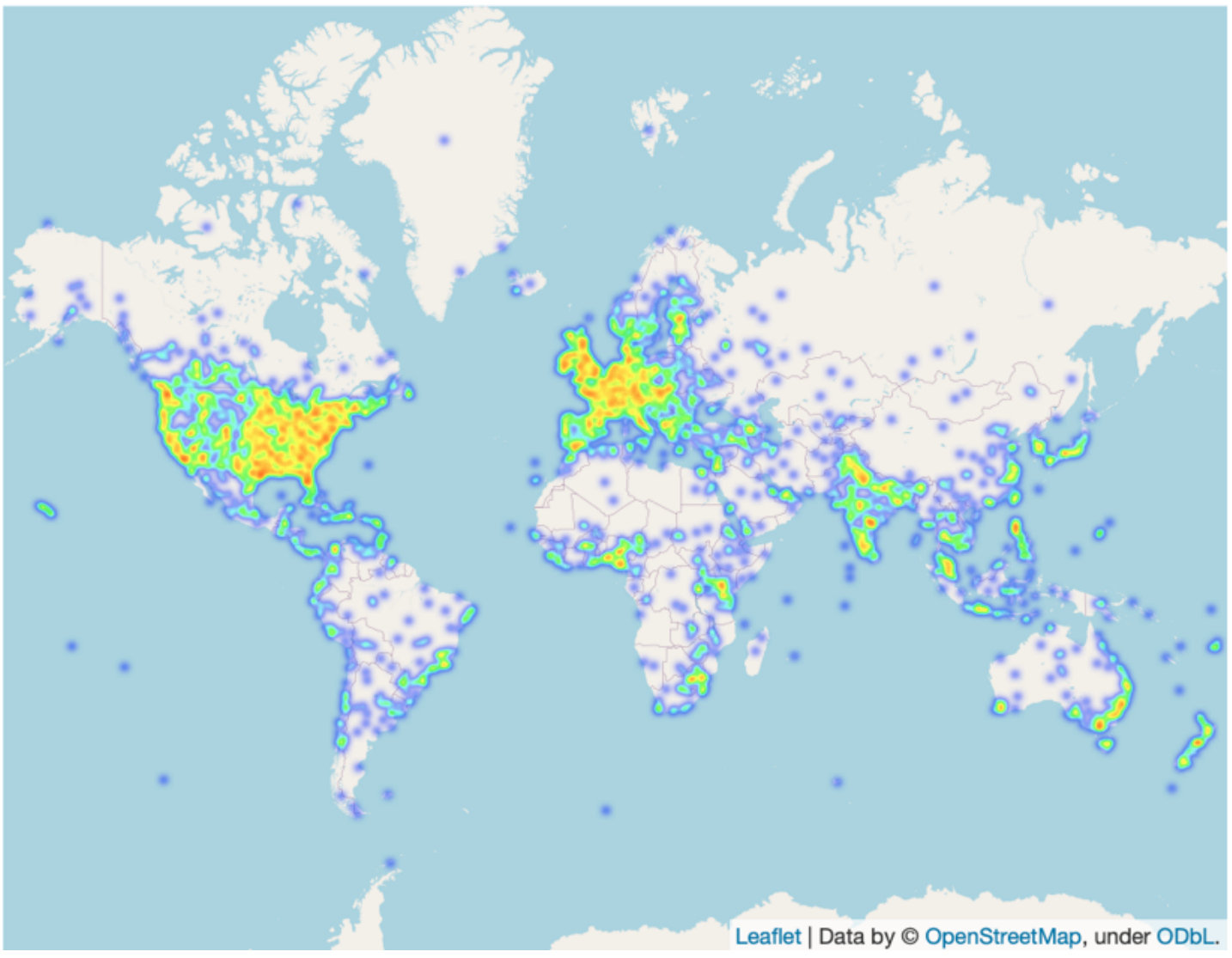

*Figure 1: Spatial Distribution of web scraped Twitter Dataset after preprocessing*

For each tweet within each dataset the following data was collected:

- Timestamp
- User ID (which was removed for privacy reasons)
- Number of Likes
- Number of Retweets
- Tweet Content
- User Defined Location (in words)

The dataset for COF was web scraped using the yfinance API which can access historical minute level data from Yahoo Finance. The following data was collected for COF (code: CL=F) in 5-minute intervals between 18/10/2020-30/10/2020:

- Timestamp
- Open Price
- High Price
- Low Price
- Close Price
- Adjusted Close Price (ACP)
- Volume

This contained 2262 datapoints which is less than the 14,400 datapoints that would be expected for 5-minute intervals over 10 days. This discrepancy is due to trading only occurring Sunday – Friday 6:00pm – 5:00pm with a 60-minute break at the beginning of each day.

### B. *Analytical Questions*

In this study, the following will be answered:

- Can Tweets about the Climate Emergency be split into distinct events using Latent Dirichlet Allocation?
- What are the general attitudes towards the Climate Emergency among Twitter Users and how does this vary?
- Is there a correlation between sentiment towards climate change and can this be used to perform one-step forward prediction and forecasting for COF?

## III. Analysis

### A. *Text Preprocessing*

The 6 CSV files obtained from the Twitter API required extensive preprocessing to prepare for analysis. This included removing rows that weren't in English manually. Tweets that were destroyed during the writing of the original CSV files were removed as well as any extra columns.

In order to prepare data for feature extraction and modelling, natural language processing was done using the NLTK package. This involved removing special symbols within the text, removing any stopwords, lemmatizing and tokenizing the text. When removing stopwords, it was decided to remove the words that made up the original queries ("Climate Change", "Climate Emergency" etc.) as these would be in all tweets and dominate any word clouds that would be produced. Lemmatization was done using the Spacy large English language library [8].

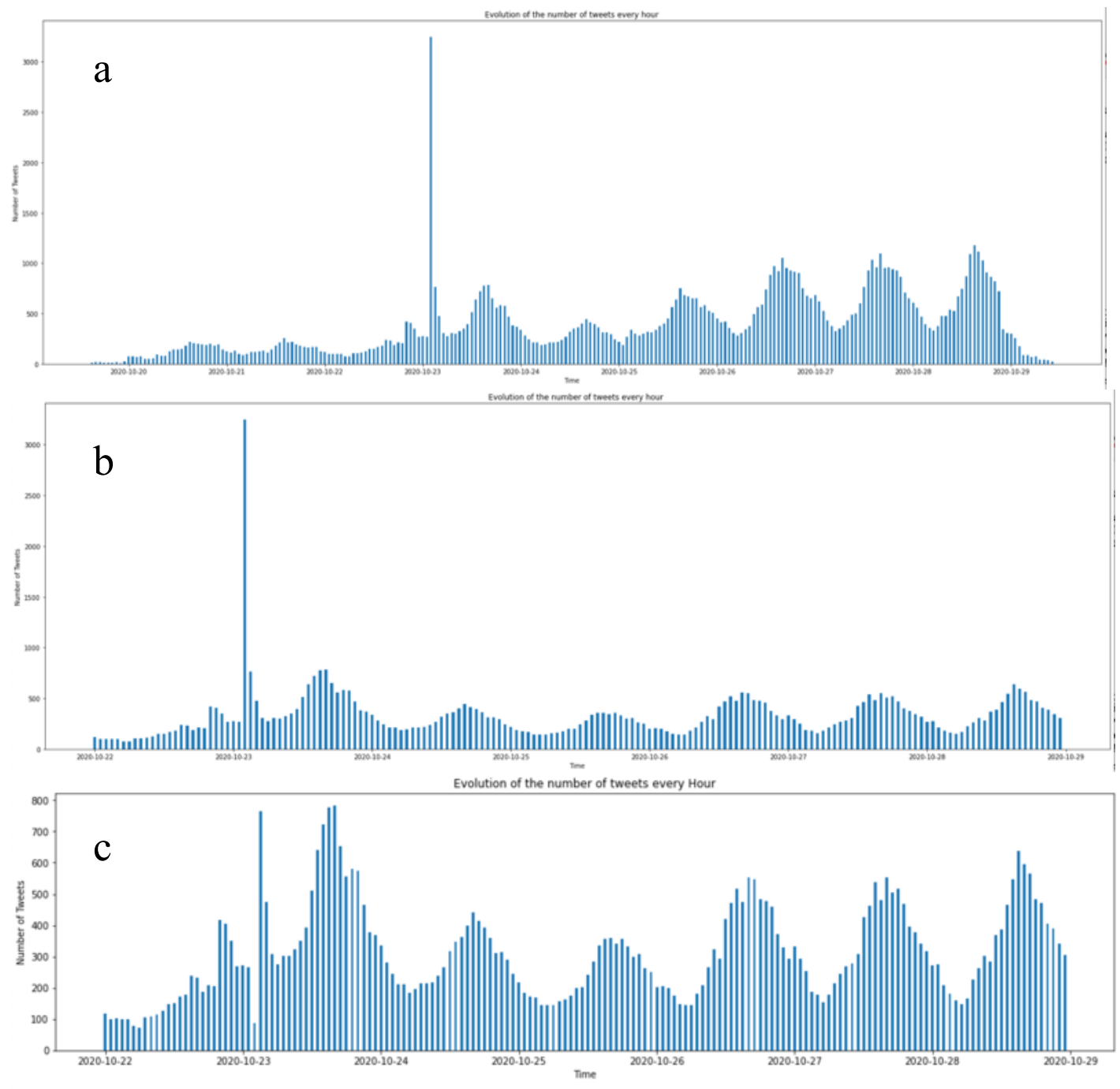

*Figure 2:a) Evolution of Tweets per hour using all search query datasets. b) Evolution of Tweets per hour after removing Climate Change dataset. c) Evolution of Tweets per Hour after outlier removal*

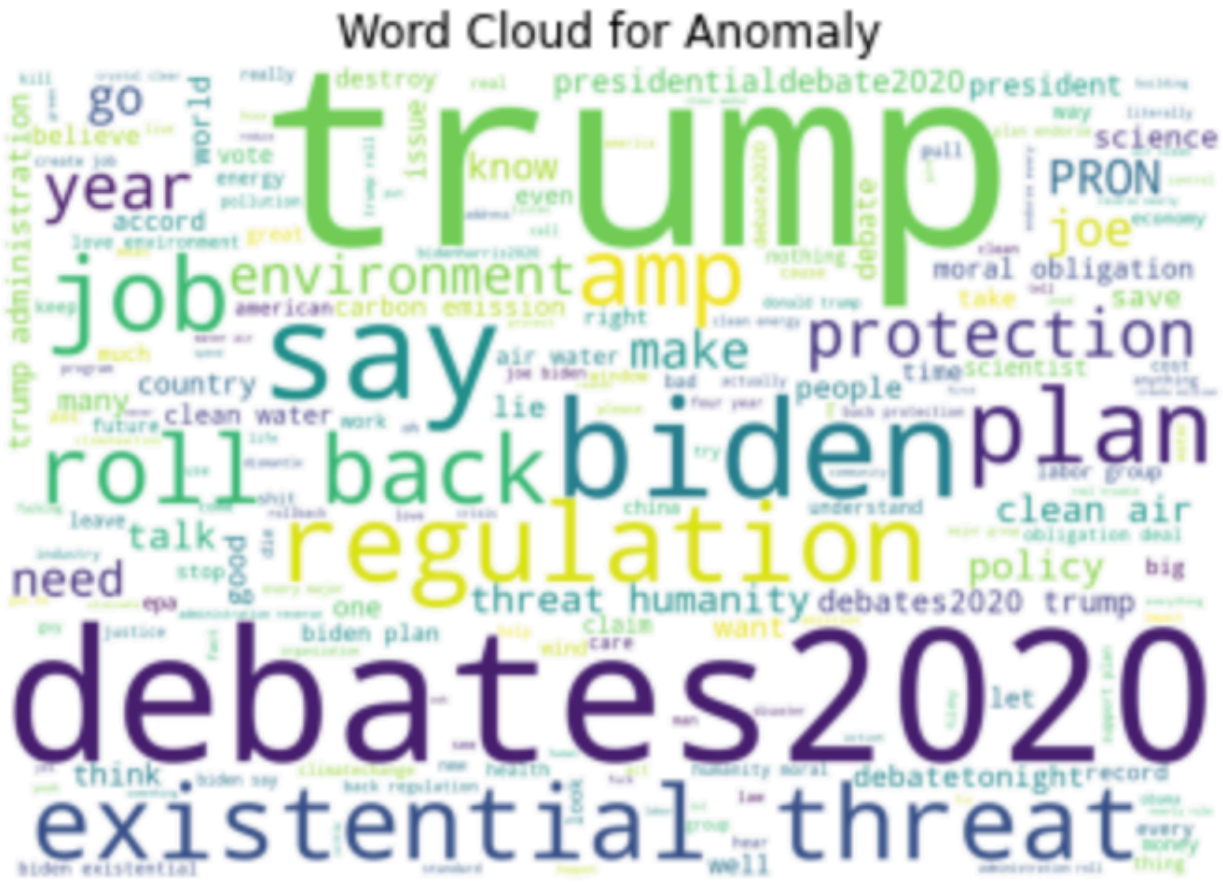

*Figure 3: Word Cloud for Outlier on 23/10/2020 3am BST*

In addition to this, data was prepared for resampling by ensuring that the index for all dataframes was the timestamp in DateTime format.

### B. *Sentiment Extraction*

In other studies sentiment analysis has been done using predictive models such as SVM, Naïve Bayes and tree ensemble methods [9] [10]. In this study, due to time constraints, this has been done using the TextBlob Rule Based Sentiment Analysis function [11]. The TextBlob library can achieve accuracy of 76% [12] for sentiment analysis of tweets. Sentiment was calculated after cleaning and preprocessing of the text data to ensure that neutral words do not have an effect on the final sentiment calculated.

During exploratory data analysis, word clouds and distribution of sentiment were plotted for tweets within each query. This was done to see whether there was a distinct difference in the topics that came up within the different dataframes.

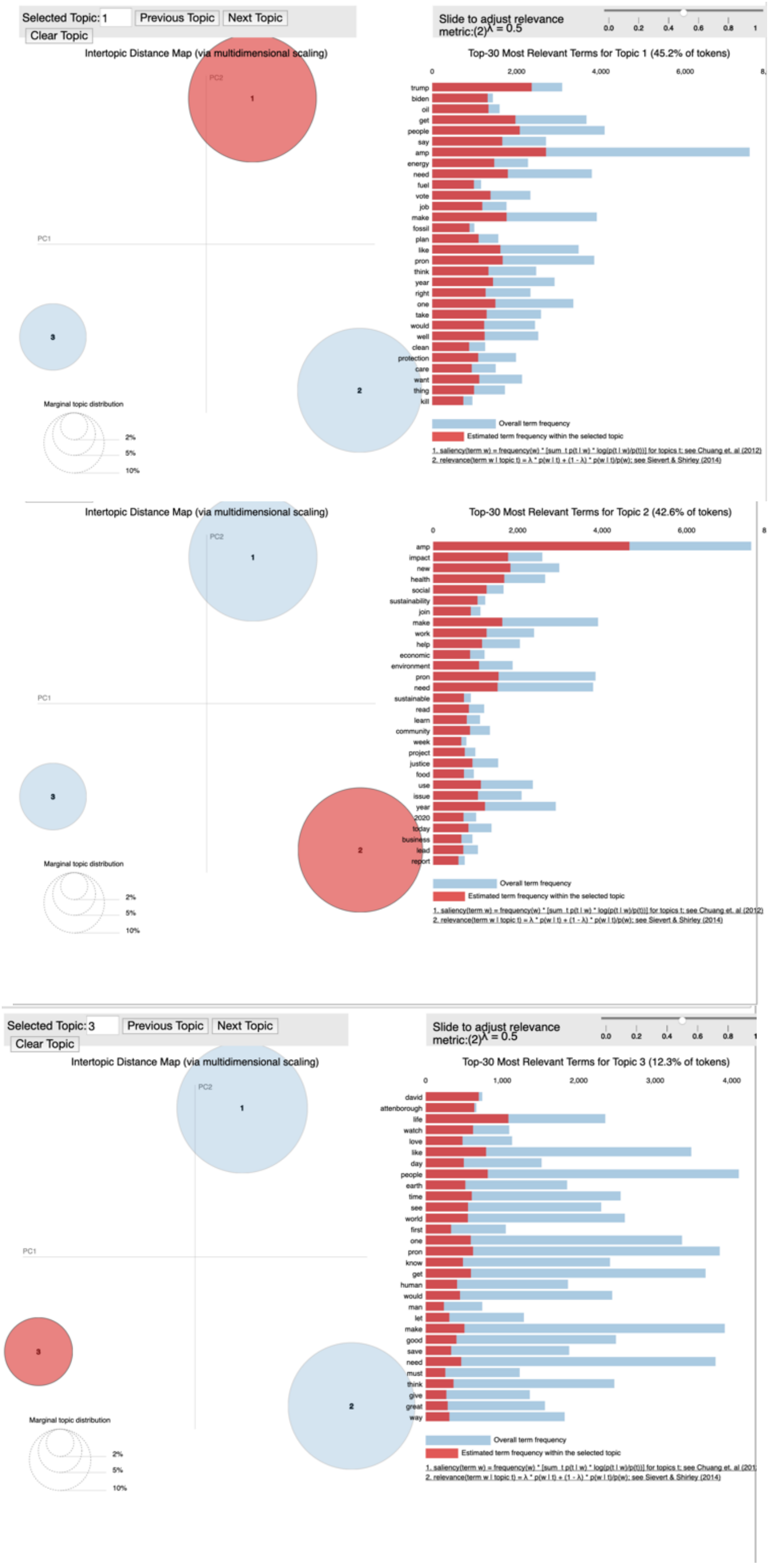

*Figure 5: Three different topics produced with keywords shown in each. Visualisation of LDA done using t-SNE in the pyLDAvis package*

## C. Data resampling and Visual Event Detection

After exploratory data analysis of the original queries, it was clear that there were distinct different topics that came up among the different tweets in the corpus. The twitter datasets were concatenated and resampled before topic modelling in line with Cheng and Wicks [14]. During this process, the following columns were created:

- Timestamp (5-minute intervals)
- Likes (sum, standard deviation and mean)
- Retweets (sum, standard deviation and mean)
- Sentiment (sum, standard deviation and mean)
- Aggregation of lemmatized, tokenized words within tweet text for each 5-minute period
- Tweet count
- Latitude (mean, standard deviation)
- Longitude (mean, standard deviation)

When plotting the tweet counts every 5 minutes for each query, there was a clear outlier on the 23/10/2020 at approximately 3am BST (Figure 2). In addition, when plotting the tweet counts for all queries between the 19/10/2020 and 29/10/2020 there was a clear skew in the data that was caused by the "Climate Change" query dataset as this was a dataset of approximately 50,000 tweets between only the 25/10/2020 and the 29/10/2020. As a result, this dataset wasn't used in this analysis.

Plotting word clouds for the outlier on the 23/10/2020 showed that this was clearly related to a political event and related to the US presidential elections which took place on the 23/10/2020 at 3am BST (Figure 3). For the purposes of this study, the largest outliers were removed from the dataset. This resulted in a final dataset with ~48,000 tweets.

## D. Unsupervised Topic Modelling using Latent Dirichlet Allocation (LDA)

The aggregated words for each interval were vectorized and used for LDA [14]. This would assign topics based on the word vector for each 5-minute interval. This is good for event detection as when there are multiple people tweeting about the same thing, there will be clearly defined groups at these times. However, within each 5-minute interval it is very possible that people are tweeting about different topics. The method that has been pursued in this study relies on the assumption that coherent topics are tweeted about in each 5-minute interval. During vectorization the minimum occurrence of each word is 10 and maximum number of features (words) is 5000 so that the model only uses the most relevant words. During LDA, a random state of 20 is used.

Using this method, 3 general topics are found after testing (Figure 4). These are:

- Politics
- Public Health

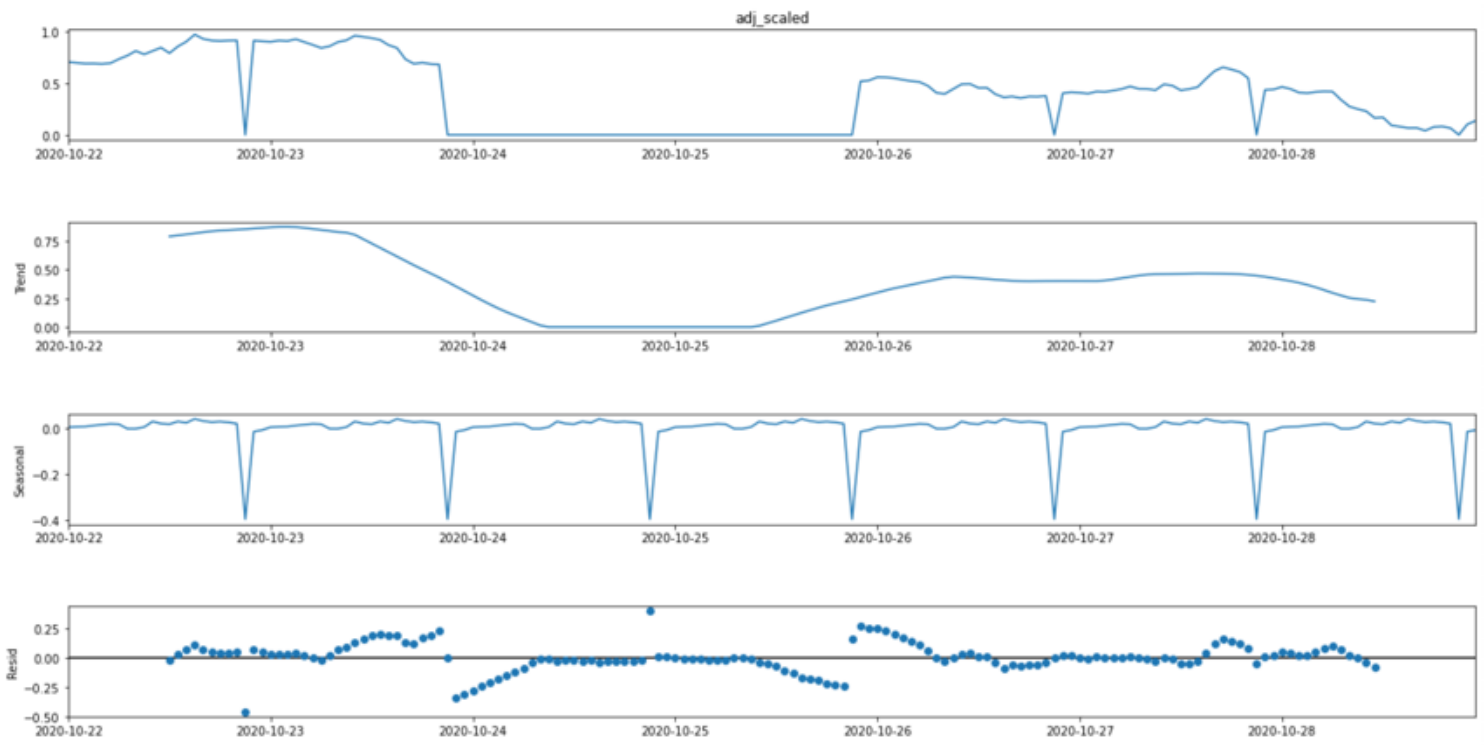

*Figure 4: Decomposition of Adjusted Close Price for Crude Oil Futures*

- The Natural World

When deciding on how many topics to split the data into, a range of 3-8 was tested. These topics have been visualized using the dimensionality reduction technique t-distributed stochastic neighbor embedding (TSNE) [15] [16] [17] .

### E. *Seasonal AutoRegressive Integrated Moving Average with Exogenous Regressors (SARIMAX) modelling*

When plotting a correlation matrix on the resampled dataset, it was clear that sum of liked and retweets were linearly related to of the tweet count. Similarly, the sum of the sentiment was a function of the tweet count. As a result, the sentiment per tweet and tweet count were used as features. Similarly, the ACP was used as a feature to represent COF.

Decomposing the different features shows the general trend, seasonal trends and residuals for each feature (Figure 5). This showed that the sentiment per tweet and the adjusted close have strong seasonal features which should be taken into account when model building.

In this study, the sentiment per tweet and tweet count all exogenous variables when predicting COF as these are considered to be not directly affected by the ACP. A Grid Search is carried out to find the optimal number of AutoRegressive (AR) parameters, differences and Moving Average (MA) parameters as well as the optimal order of the season component for these respective parameters. Performance of each model during grid search is measured using the Akaike Information Criterion (AIC) which is a measure of out of sample error [19].

SARIMAX models are a popular linear model for forecasting seasonal time series data [20][21]. In this case it will be used to predict the short term (next hour) ACP.

### F. *Validation of SARIMAX Model Results*

Before modelling, data is split into training and testing data in a 70%:30% ratio. When evaluating the model on the test data, the root mean squared error (RMSE) between the true values of the ACP and the predicted values. This is compared with the training set error as a form of validation.

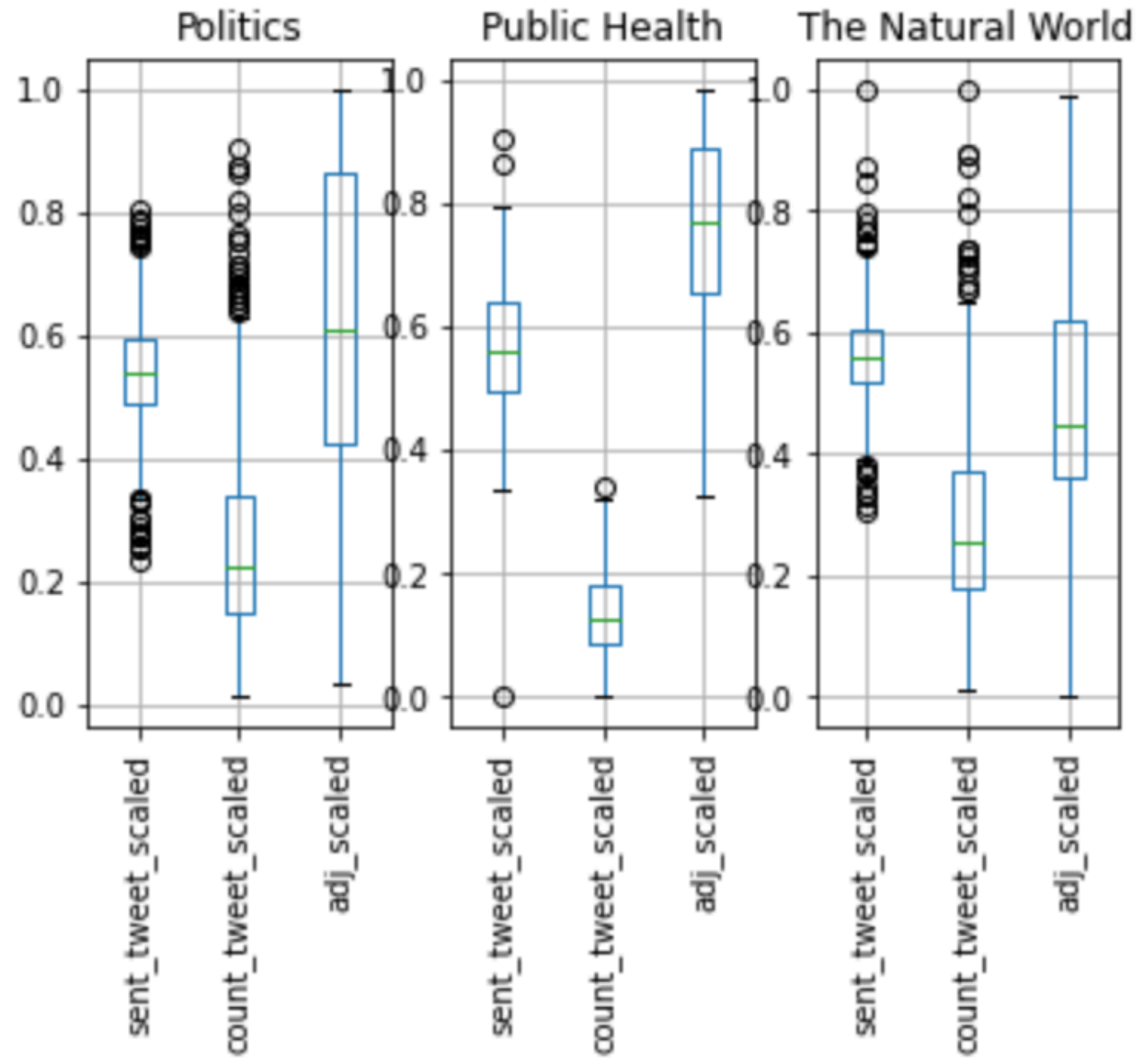

*Figure 7: Boxplots to show the sentiment per tweet, tweet count and adjusted close price*

## IV. Findings, Reflections and Further Work

### A. *Event Detection using LDA*

It was possible to split the tweets into distinct topics using LDA and t-SNE to visualize the results. Using this, it is possible to detect particular events within a corpus of tweets (Figure 6). The data shows patterns in time where different topics are discussed on twitter.

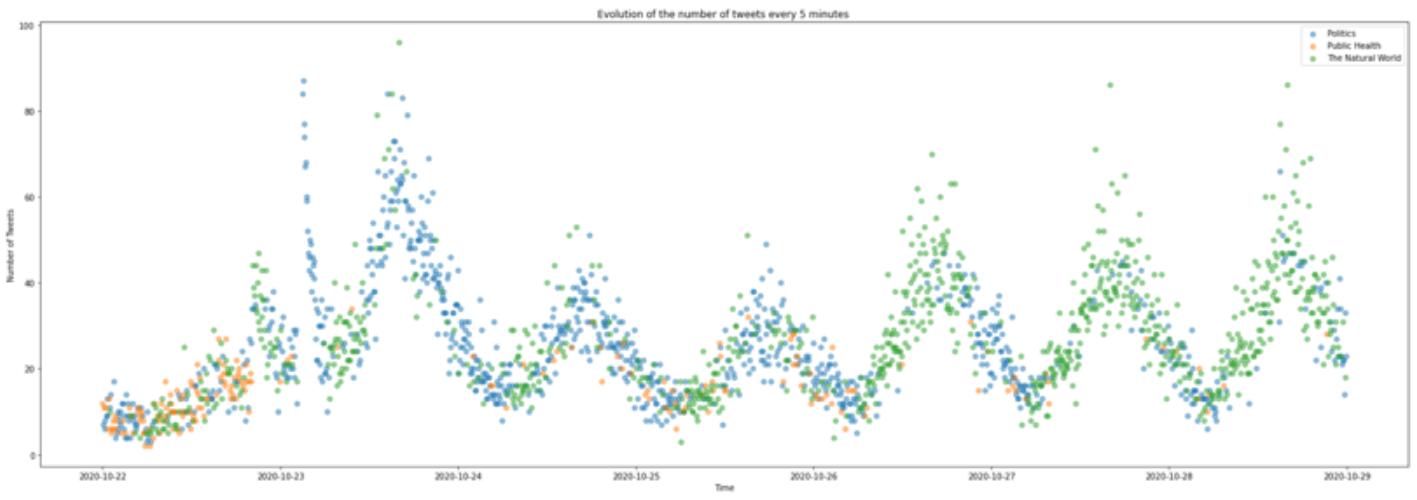

*Figure6: Evolution of number of tweets every 5 minutes for each topic*

### B. *Variation in Attitudes towards the Climate Emergency*

The results from this dataset surrounding variation in sentiment for each topic are inconclusive. The distributions of the sentiment per tweet (Figure 7) are very similar with the Public Health topic showing positive skewness (Figure 8) and a very similar mean and standard deviation to The Natural World topic. The political topic has a lower mean sentiment per tweet with a smaller standard deviation which could show generally more negative sentiment for this category.

### C. *Forecasting Crude Oil Futures using SARIMAX modelling*

The model produced has a training RMSE of 0.196 and a testing RMSE of 0.209 using this dataset. When observing the one-step forward prediction (Figure 9), the model is able to capture increases and decreases in the observed data on the 23/10/20 and 26/10/20. However, the changes that are predicted are sharp in their nature. This would be due to the sharp increases in tweet count at these times which would encourage the model to predict sharp changes in the COF Adjusted Close.

When carrying out multiple step forecasting, the model is unable to capture changes in price that deviate from the trend of the training data. This would be due to the MA and AR parameters that were chosen that follow the trend of the training data closely. The forecast does not track increases and decreases in data as well as the one step forward prediction can. This indicates that using tweet count and sentiment per tweet as exogenous variables in a SARIMAX model can be used to predict short term changes (hour by hour) but not long term changes effectively.

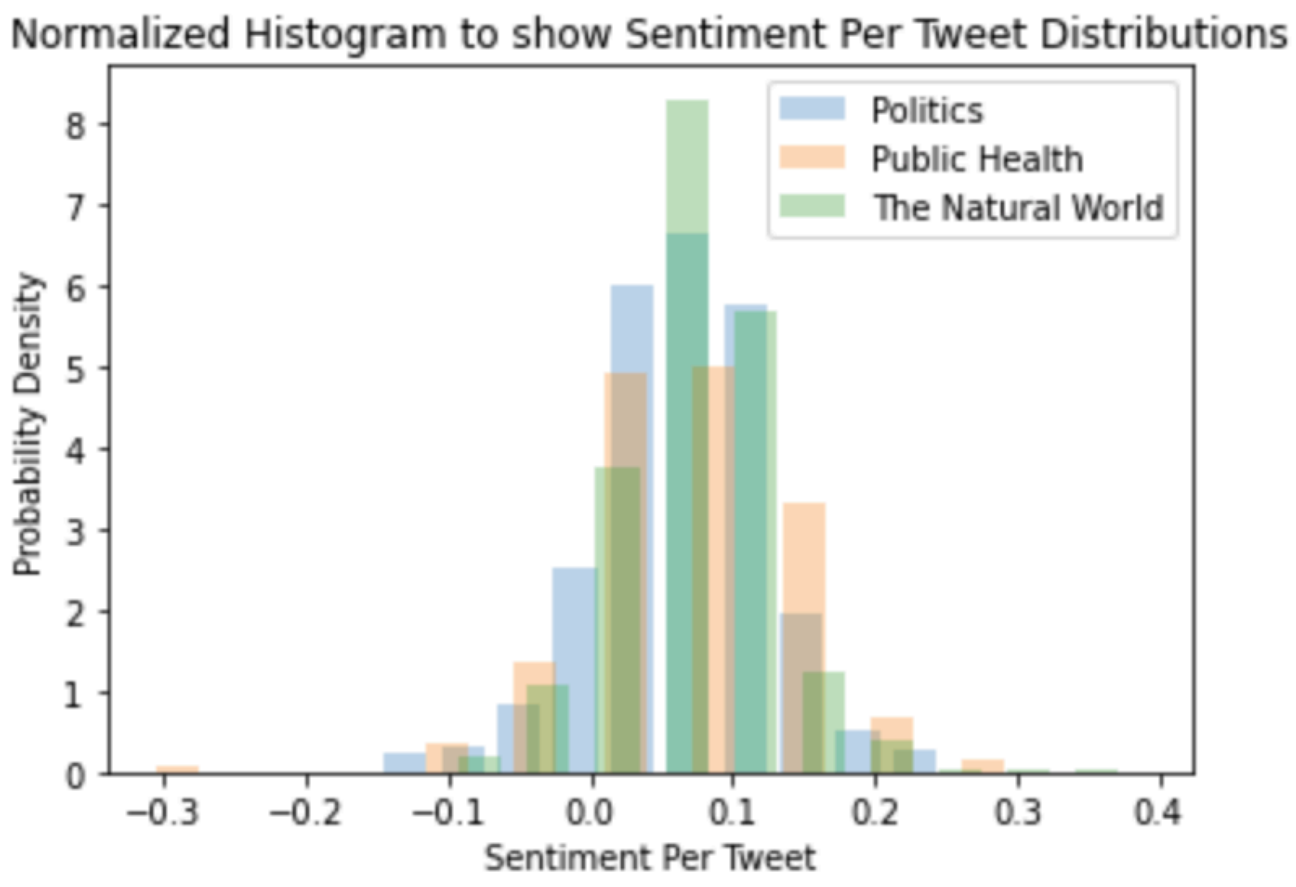

*Figure 8: Normalized Histogram to show Sentiment Per Tweet Distributions*

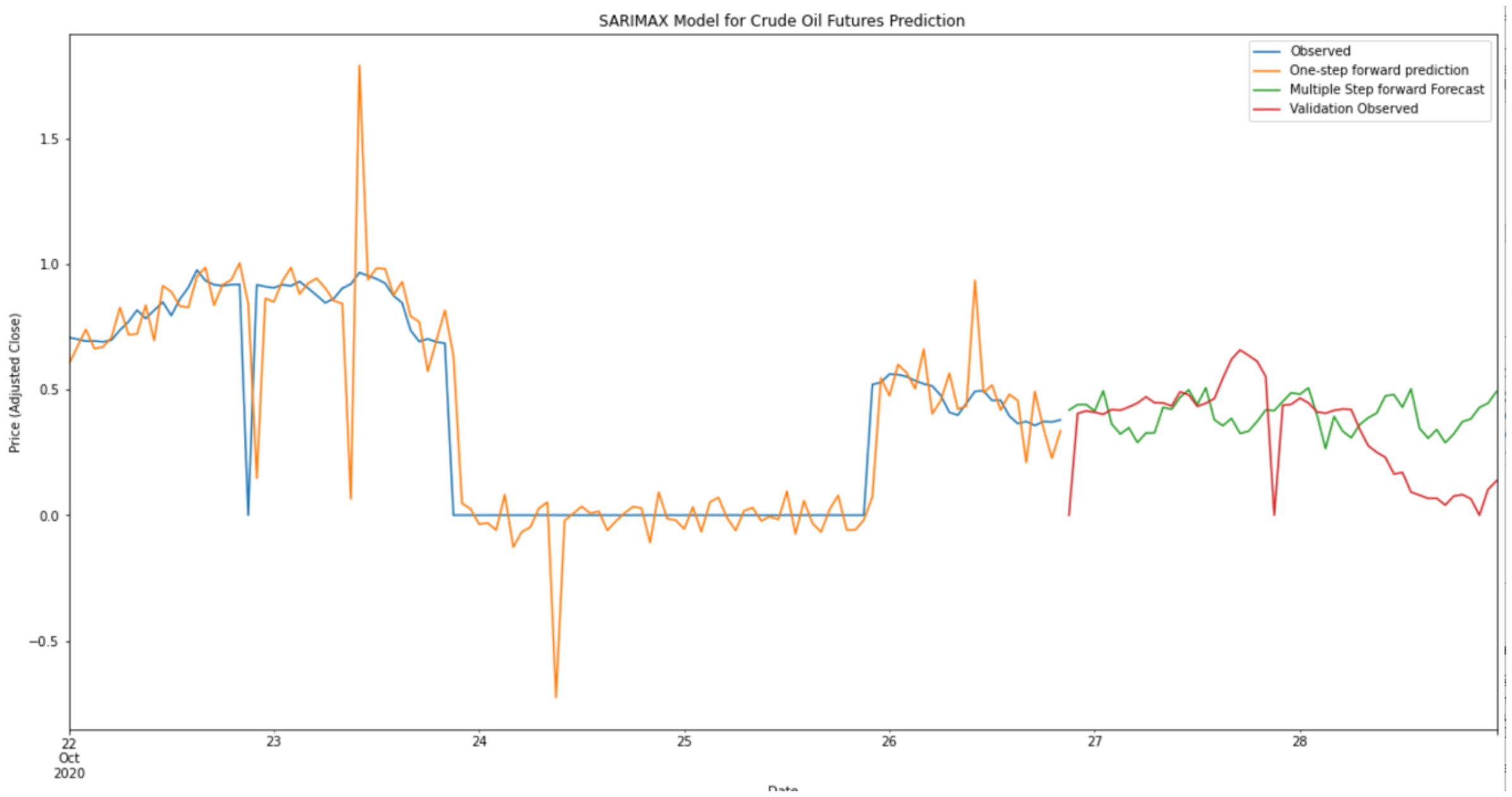

*Figure 9: SARIMAX Model for Crude Oil Futures Forecast and One Step Forward Prediction*

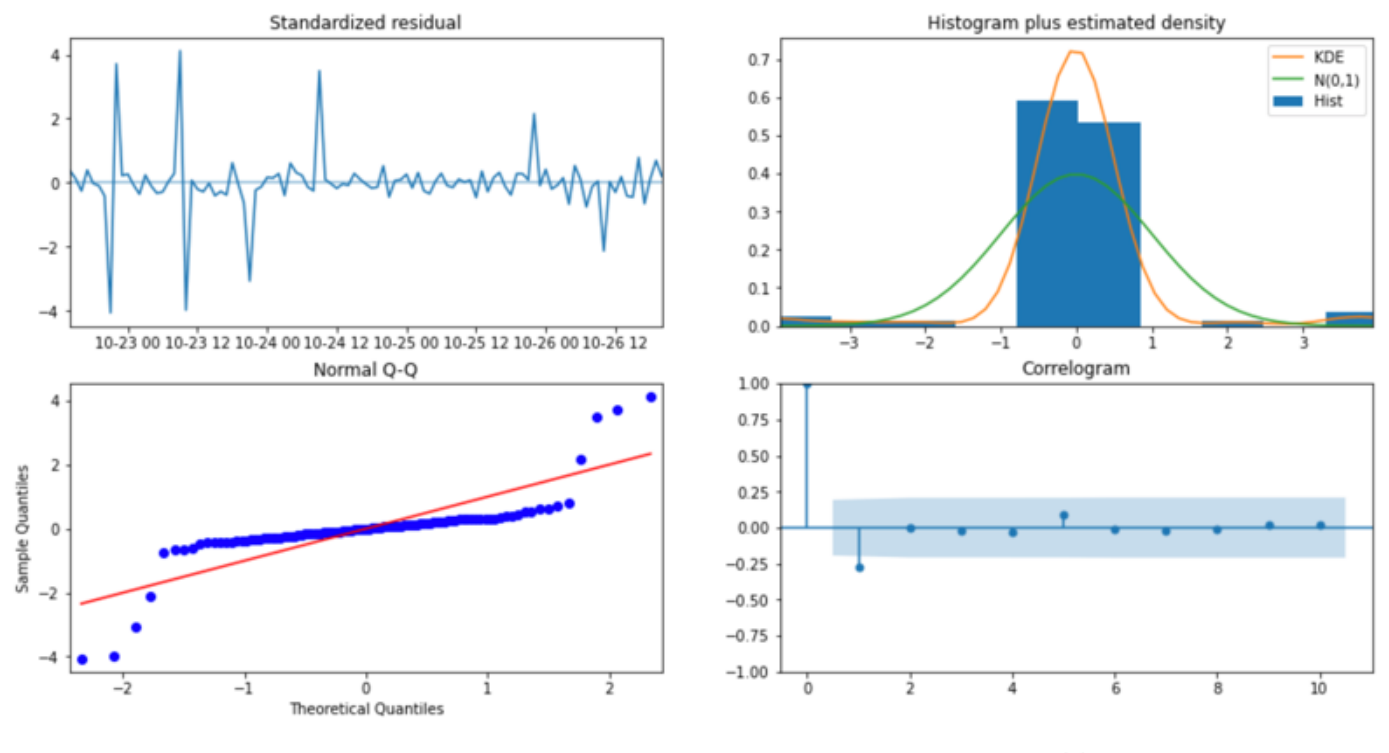

*Figure 10: Diagnostic statistics from SARIMAX model plotted using statsmodel API 'plot_diagnostic' function*

## D. Reflections

Through the building of an unsupervised topic model, it could be argued that this study has identified the most important issues surrounding the climate emergency. The topic with the highest sentiment was Public Health. It could be argued that policy which puts public health first in the context of climate, such as depollution initiatives, could be a means to effective action.

These results show that there are no linear relationships between the two but that tweet count and sentiment per tweet can be used as exogenous variables in one-step forwards prediction. The Public Health topic had the largest correlation to the ACP for Crude Oil Futures. This could be explained by individuals having the strongest reaction to the climate emergency when it is seen to cause personal health problems

## E. Further Work

To further this work, the spatial-temporal properties of sentiment towards the climate emergency could be interrogated. This would be done with density-based clustering techniques, such as DBSCAN. This was done with this dataset but due to time constraints of the project and the long run time of Scikit-Learn's DBSCAN algorithm, conclusive results could not be found.

Modelling using the SARIMAX model was effective, but it would be interesting to see how whether the use of more complicated long-short term memory (LSTM) models could yield better results [22]. Predicting increases and decreases in ACP may be more appropriate that predicting the ACP itself.

Lastly, increasing the scale of the project could provide more conclusive results. A time period of 7 days makes it very difficult to capture long standing patterns in human behavior on Twitter. In order to compare with other literature, the study must take place over a period of months and years. However, this would make processing more computationally expensive.

## VI. Word Count

| | |
|---|---|
| Abstract | 144 |
| Introduction | 282 |
| Analytical Questions and Data | 328 |
| Analysis | 1012 |
| Findings, reflections and further work | 614 |